\documentclass{article}
\usepackage{spconf,amsmath,graphicx,hyperref}
\usepackage{amssymb}

\title{DEMO: Disentangled Motion Latent Flow Matching for Fine-Grained Controllable Talking Portrait Synthesis }
\name{Peiyin Chen\sthanks{This work is supported by the Fundamental Research Funds for the Central Universities (B250201085) and Changzhou Science and Technology Project (CJ20240093).}$^1$, Zhuowei Yang$^2$, Hui Feng$^1$, Sheng Jiang$^3$, Rui Yan\sthanks{represents corresponding author.}$^4$}

\address{$^1$College of Artificial Intelligence and Automation, Hohai University, Changzhou, China \\
         $^2$College of DaYu, Hohai University, Nanjin, China \\
         $^3$College of Water Conserwancy and Hydropower Engineering, Hohai University, Nanjing, China\\
         $^4$College of Computer Science and Technology, Zhejiang University of Technology, Hangzhou, China}

\usepackage{setspace}
\usepackage{etoolbox}
\AtBeginEnvironment{thebibliography}{%
  \setstretch{0.9}
  \setlength{\itemsep}{0pt}%
  \setlength{\parskip}{0pt}%
  \setlength{\parsep}{0pt}%
}

\begin{document}
%
\maketitle
\begin{abstract}
Audio-driven talking-head generation has advanced rapidly with diffusion-based generative models, yet producing temporally coherent videos with fine-grained motion control remains challenging. We propose DEMO, a flow-matching generative framework for audio-driven talking-portrait video synthesis that delivers disentangled, high-fidelity control of lip motion, head pose, and eye gaze. The core contribution is a motion auto-encoder that builds a structured latent space in which motion factors are independently represented and approximately orthogonalized. On this disentangled motion space, we apply optimal-transport–based flow matching with a transformer predictor to generate temporally smooth motion trajectories conditioned on audio. Extensive experiments across multiple benchmarks show that DEMO outperforms prior methods in video realism, lip–audio synchronization, and motion fidelity. These results demonstrate that combining fine-grained motion disentanglement with flow-based generative modeling provides a powerful new paradigm for controllable talking-head video synthesis.

\end{abstract}
\section{Introduction}
\label{sec:intro}
\hspace{1em}Portrait animation, or talking-head generation, aims to synthesize dynamic facial videos from a single static image conditioned on audio. It supports applications in film production, virtual communication, and interactive gaming, where accurate lip synchronization, natural head motion, and expressive eye gaze are essential for immersive human–computer interaction. Despite recent progress, audio-driven portrait animation remains challenging because speech and facial motion exhibit an inherent one-to-many relationship: the same utterance can correspond to diverse expressions, head poses, and gaze patterns, which makes it difficult to generate motion that is both temporally precise and semantically coherent using audio alone.

\hspace{1em}Recent diffusion-based generative models, including Stable Diffusion \cite{rombach2022high}, DiT \cite{peebles2023scalable} and flow-matching methods \cite{ki2024float}, have substantially improved image and video synthesis by injecting noise into latent representations and learning to invert this process to produce highly realistic and diverse results. In addition, parametric and implicit representations of lip motion \cite{radford2021learning}, facial expressions \cite{molad2023dreamix} and head pose, when combined with latent-space diffusion \cite{luo2023videofusion}, partly reduce the ambiguity in audio-to-motion mapping. However, existing methods still lack fine-grained, disentangled control over motion factors, leading to entanglement of lips, eyes, and head movements, and they often produce noisy, temporally inconsistent trajectories with limited computational efficiency. Consequently, they struggle to control factors such as eye gaze or require simultaneous modification of all motions, constraining both flexibility and practical applicability.

\hspace{1em}To address these challenges, we propose DEMO, an audio-driven talking-portrait video generation framework based on flow-matching generative modeling. DEMO employs a motion auto-encoder that learns a structured, fine-grained latent space where lip motion, head pose, and eye gaze are disentangled and approximately orthogonalized, enabling precise and independent control of each motion factor. On this latent representation, we apply optimal-transport flow matching with a transformer-based vector-field predictor to efficiently generate audio-conditioned motion trajectories with strong temporal coherence. Our main contributions in this work are: 
\begin{itemize}
\item[$\bullet$] We design a motion auto-encoder that provides a disentangled latent space for flexible and precise manipulation of facial dynamics.
\end{itemize}
\begin{itemize}
\item[$\bullet$] We propose an optimal-transport flow-matching approach with a transformer predictor for efficient, temporally consistent audio-driven motion synthesis.
\end{itemize}
\begin{itemize}
\item[$\bullet$] DEMO achieves the state-of-the-art performance in video realism, lip–audio synchronization, and motion fidelity, significantly surpassing existing methods.
\end{itemize}

\begin{figure*}[htbp]
    \centering
    \includegraphics[width=\textwidth]{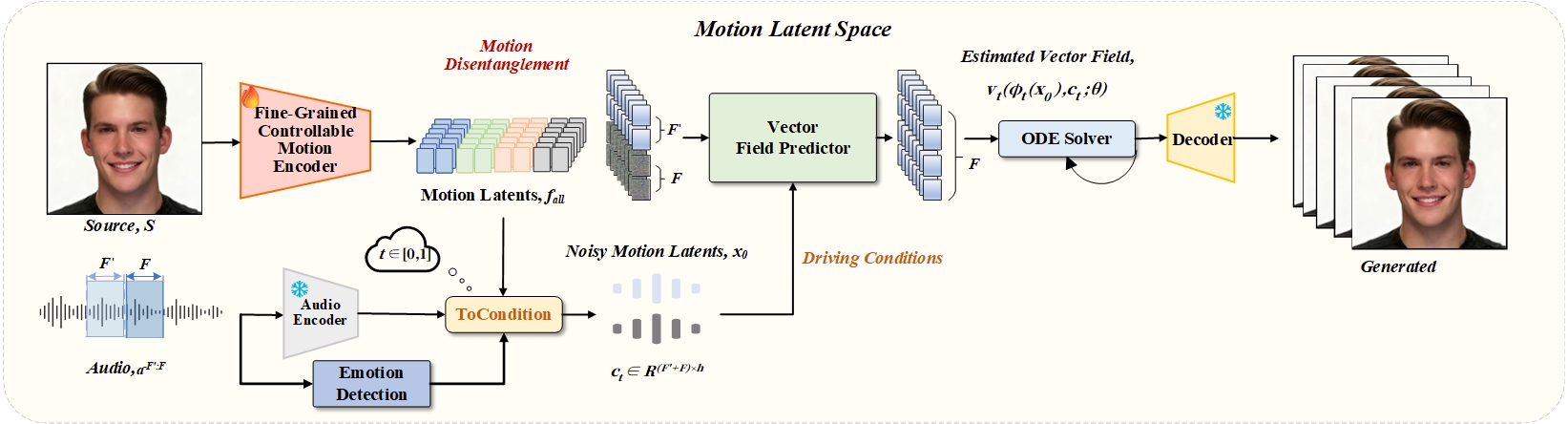}
    \caption{Overview of the proposed DEMO framework for talking-head video generation. Given a source image (left) and a driving audio sequence, DEMO employs a Fine-Grained Controllable Motion Encoder (orange) to construct a disentangled motion representation that separates lip, head-pose, and eye movements. Audio embeddings enriched with emotion cues (blue) drive the motion evolution. A Vector Field Predictor with OT-based flow matching (green) refines noisy motion latents into temporally coherent trajectories, which are integrated by an ODE solver and finally decoded into high-fidelity, synchronized video frames (right).}
    \label{fig:demo-pipeline}
\end{figure*}

\section{Method}
\label{sec:format}
\hspace{1em}We present an overview of \textbf{DEMO} in Fig.~\ref{fig:demo-pipeline}. Given a source image $S\in \mathbb{R}^{3\times H\times W}$ and a driving audio sequence $a^{1:F}\in \mathbb{R}^{F\times d^a}$, our framework generates F-frame talking head videos with synchronized verbal and non-verbal motions. DEMO operates in two stages: (1) pretraining a motion auto-encoder to construct a fine-grained latent space that enables controllable facial motion representation, and (2) applying optimal-transport flow matching \cite{lipman2022flow} with a transformer-based predictor to map audio inputs to motion latents, which are then decoded into high-fidelity video frames.

\begin{figure}[htbp]
    \centering
    \includegraphics[width=1\linewidth]{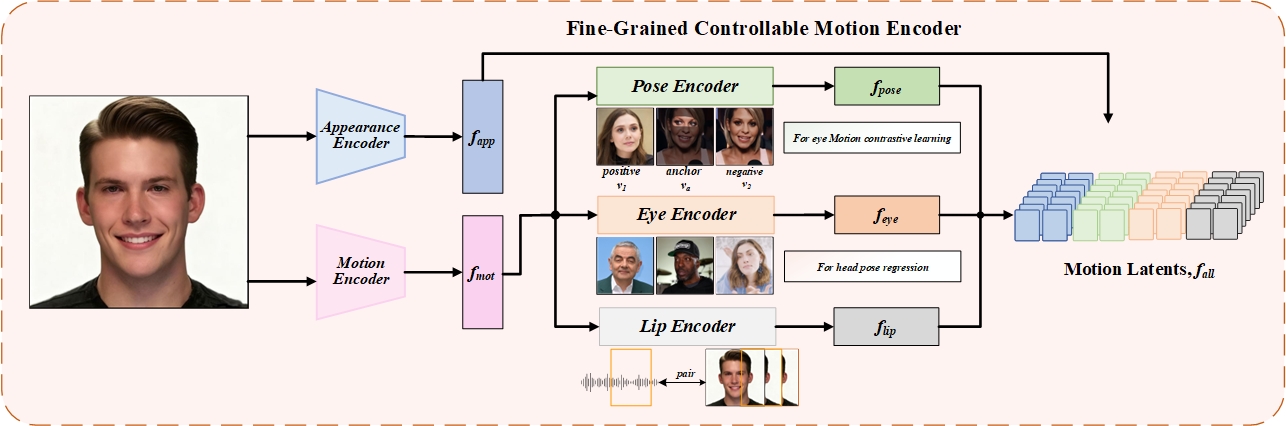}
    \caption{The structure of our Fine-Grained Controllable Motion Encoder. }
    \label{fig:motion encoder}
\end{figure}

\begin{table*}[t]
	\centering
	\renewcommand{\arraystretch}{1}
	\setlength{\tabcolsep}{11pt}
	\caption{The quantitative comparisons with the existing portrait image animation approaches on the HDTF. The best result for each metric is in bold.}
	\begin{tabular}{l|ccccc|c}
		\hline
		\textbf{Method} & 
		\multicolumn{5}{c|}{\textbf{Video Generation}} & 
		\textbf{Lip Synchronization} \\
		& FID $\downarrow$ & FVD $\downarrow$ & SSIM $\uparrow$ & CSIM $\uparrow$ & P-FID $\downarrow$ & LSE-D $\downarrow$ \\
		\hline
		Hallo (\textit{CVPR, 2024})     & 100.255 & \textbf{126.242} & 0.307 & 0.682  & 0.946 & 258.228 \\
		EDTalk (\textit{ECCV, 2025})    & 101.543 & 130.119 & \textbf{0.321} & 0.661  & 1.145 & 240.105 \\
		EchoMimic (\textit{AAAI, 2024}) & 109.331 & 142.727 & 0.306 & 0.671  & 0.985 & 264.711 \\
		SadTalker (\textit{CVPR, 2023}) & 117.746 & 157.569 & 0.315 & 0.694  & 0.642 & 302.022 \\
		\textbf{DEMO (\textit{Ours})} & \textbf{94.050} & 132.161 & 0.314 & \textbf{0.704} & \textbf{0.587} & \textbf{238.577} \\
		\hline
	\end{tabular}
	\label{tab:quantitative-comparison}
\end{table*}

\textit{A. Fine-Grained Controllable Motion Motion-Encoder} 

\hspace{1em}
Given an arbitrary person image, our goal is to synthesize a talking-head video in which facial motions such as lip movement, head pose, and eye gaze can be independently controlled. To this end, we disentangle latent visual representations in a coarse-to-fine manner to construct a fine-grained motion latent space, as illustrated in Fig.~\ref{fig:motion encoder}. We first separate appearance from motion to obtain a unified motion representation capturing all dynamic information, and then employ motion-specific contrastive learning to further disentangle individual motion components, excluding expressions, within this representation.

\hspace{1em}Concretely, \textbf{an appearance encoder} $E_{app}$ and \textbf{a motion encoder} $E_{mot}$ are employed to extract features from an appearance image and a driving frame, respectively. A generator $G_0$ synthesizes a face image with the identity of the appearance image and the motion of the driving frame. To enhance the accuracy of the extracted motion features, we introduce a motion reconstruction loss \cite{burkov2020neural}:
\begin{equation}
\mathcal{L}_{mot} = \| \phi(I_0) - \phi(I_g) \|_2^2 + \| \psi(I_0) - \psi(I_g) \|_2^2,
\end{equation}
where $\phi(\cdot)$ and $\psi(\cdot)$ are features extracted by the 3D face reconstruction and emotion networks \cite{danvevcek2022emoca}, $I_0$ is the generated image, and $I_g$ is the ground truth.

\hspace{1em}Building upon the unified motion feature, we further extract fine-grained components. For eye motion, 
we create an anchor frame by compositing the eye region from one driving frame with the remaining regions of another.
Given two driving frames $v_1$ and $v_2$, an anchor frame $v_a$ is formed by combining the eye region of $v_1$ with the other regions of $v_2$. \textbf{The eye encoder $E_{eye}$} then extracts features $(f_1, f_2, f_a)$, from which a positive pair $(f_1, f_a)$ and a negative pair $(f_2, f_a)$ are constructed. The encoder is trained to isolate eye-specific features through a contrastive loss:
\begin{equation}
\mathcal{L}_{eye} = - \log \frac{\exp(\mathcal{S}(f_1,f_a))}{\exp(\mathcal{S}(f_1,f_a))+\exp(\mathcal{S}(f_2,f_a))},
\end{equation}
where $\mathcal{S}(\cdot,\cdot)$ denotes cosine similarity.

\hspace{1em}Head pose is parameterized by three Euler angles and three translations. \textbf{A pose encoder} $E_{pose}$ directly regresses these parameters under the supervision of a 3D face prior:
\begin{equation}
\mathcal{L}_{pose} = \|P_{pred} - P_{gt}\|_1.
\end{equation}
\hspace{1em}Finally, lip motion is modeled using audio-visual contrastive learning \cite{zhou2021pose}. A lip encoder $E_{lip}$ and an audio encoder $E_{aud}$ extract motion features $f_i^v = E_{lip} \circ E_{mot}(v_i)$ and audio features $f_i^a = E_{aud}(a_i)$. Positive and negative audio-visual pairs are constructed to enforce consistency via InfoNCE losses \cite{oord2018representation}:
\begin{equation}
\mathcal{L}_{a2v} = -\log \frac{\exp(\mathcal{S}(f_i^a,f_i^v))}{\exp(\mathcal{S}(f_i^a,f_i^v))+\sum_{k=1}^K \exp(\mathcal{S}(f_i^a,f_k^v))},
\end{equation}

\begin{equation}
\mathcal{L}_{v2a} = -\log \frac{\exp(\mathcal{S}(f_i^v,f_i^a))}{\exp(\mathcal{S}(f_i^v,f_i^a))+\sum_{k=1}^K \exp(\mathcal{S}(f_i^v,f_k^a))},
\end{equation}
\hspace{1em}This ensures that lip motion features from video and audio remain well aligned, completing the disentanglement of fine-grained controllable motions. By jointly isolating eye gaze, head pose, and lip dynamics within a linear and approximately orthogonal latent space, the auto-encoder yields a structured representation of motion. With this motion space, we perform optimal-transport–based flow matching to sample temporally consistent motion trajectories, as detailed in the next section.

\vspace{\baselineskip}

\textit{B. Flow Matching in Motion Latent Space}

\hspace{1em}With the disentangled and approximately orthogonal motion space, we employ \textbf{OT-based flow matching} \cite{paszke1912imperative} to sample motion trajectories. Specifically, we predict a vector field $\mathbf{v}_t(x_t,c_t;\theta) \in \mathbb{R}^{F\times d}$, where $x_t$ is the sample at flow time $t \in [0,1]$, and $c_t \in \mathbb{R}^{F\times h}$ denotes the driving conditions for $F$ consecutive frames. By solving the corresponding ODE, this vector field defines a flow $\varphi_t : [0,1] \times \mathbb{R}^{F\times d} \rightarrow \mathbb{R}^{F\times d}$, which produces temporally coherent motion latents.

\hspace{1em}Our vector field predictor is built on the transformer encoder \cite{zhu2018arbitrary} following the DiT \cite{sun2022landmarkgan}. Unlike DiT, where all tokens are modulated by a shared diffusion timestep and class embedding through adaptive layer normalization (AdaLN), we separate frame-wise conditioning from temporal modeling. Each frame latent is first modulated by its own condition embedding, and temporal dependencies are then captured with masked self-attention over $2\cdot T$ neighboring frames to ensure consistent motion dynamics across time. Formally, for the $f$-th frame at flow time $t$, frame-wise AdaLN and gating are applied as:

\begin{equation}
\gamma_i^f \cdot \mathrm{LN}(X_t^f) + \beta_i^f \in \mathbb{R}^h, \quad 
\alpha_i^f \cdot X_t^f \in \mathbb{R}^h,
\end{equation}

where $i \in \{1,2\}$, $h$ is the hidden dimension, and $X_t^f$ denotes the input latent of the $f$-th frame. The modulation coefficients $\alpha_i^f, \beta_i^f, \gamma_i^f \in \mathbb{R}^h$ are produced from the condition $c_t^f$ through a linear layer.

\section{Experiments}
\label{sec:pagestyle}

\textit{A. Experiments}

\hspace{1em}We train the motion encoder on three datasets: MEAD \cite{wang2020mead}, RAVDESS \cite{livingstone2018ryerson}, and HDTF \cite{zhang2021flow}. MEAD contains over 300 identities, RAVDESS provides 2400 emotion-rich clips from 24 speakers, and HDTF offers broader identity diversity. All videos are converted to 25 FPS, audio is resampled to 16 kHz, and cropped faces are resized to 512×512 following \cite{siarohin2019first}. For HDTF, we use 6.9 hours of 5000 clips from 4600 identities for training and 400 unseen identities for testing. For RAVDESS, 22 identities are used for training and 2 for testing, with non-overlapping splits across datasets.

\hspace{1em}The motion latent dimension is set to 512. The vector predictor adopts a Transformer with 8 attention heads and a hidden size of 1024. Input sequences consist of 50 frames with 10 preceding frames. Training employs the Adam \cite{kinga2015method} with batch size of 16, learning rate of $10^{-4}$, L1 loss, and balancing coefficients $\lambda_{\text{OT}}=0.6, \lambda_{\text{vel}}=1$. The model is trained for 20k steps ($\approx$2 days) on two NVIDIA A100 GPUs, using the Euler method \cite{lipman2022flow} as the ODE solver.

\vspace{\baselineskip}

\begin{table*}[htbp]
	\centering
	\renewcommand{\arraystretch}{1} 
	\setlength{\tabcolsep}{16pt}     
	\caption{Ablation studies of DEMO on HDTF dataset. The best result for each metric is in bold.}
	\begin{tabular}{l|ccccc|c}
		\hline
		\textbf{Method} & 
		\multicolumn{5}{c|}{\textbf{Video Generation}} & 
		\textbf{Lip Synchronization} \\
		& FID $\downarrow$ & FVD $\downarrow$ & SSIM $\uparrow$ & CSIM $\uparrow$ & P-FID $\downarrow$ & LSE-D $\downarrow$ \\
		\hline
		VAE+Flow   & 121.311 & 187.357 & \textbf{0.318} & 0.678 & 0.723 & 243.227 \\
		FCME+Diff  & 118.966 & 177.368 & 0.282 & 0.674 & 1.543 & 246.487 \\
		FCME+Flow  & \textbf{94.050} & \textbf{132.161} & 0.314 & \textbf{0.704} & \textbf{0.587} & \textbf{238.577} \\
		\hline
	\end{tabular}
	\label{tab:ablation}
\end{table*}

\textit{B. Evaluation Metrics and Baselines}

\hspace{1em}To comprehensively assess both image and video quality, we use Fréchet Inception Distance (FID) \cite{seitzer2020pytorch} to evaluate frame-level realism and Fréchet Video Distance (FVD-16) \cite{unterthiner2018towards} to measure temporal coherence across 16-frame sequences. 
Motion fidelity is evaluated with Cosine Similarity of identity embeddings (CSIM) \cite{deng2019arcface} for identity preservation, Expression FID (E-FID) \cite{tian2024emo} for expression accuracy, and Pose FID (P-FID) for head-pose consistency. For audio–visual alignment, we further report Lip-Sync Error Distance (LSE-D) and Lip-Sync Error Confidence (LSE-C) \cite{prajwal2020lip}. Together, these metrics provide a balanced evaluation of perceptual quality, motion fidelity, and synchronization precision.

\begin{figure}[htbp]
	\centering
	\includegraphics[width=0.5\textwidth]{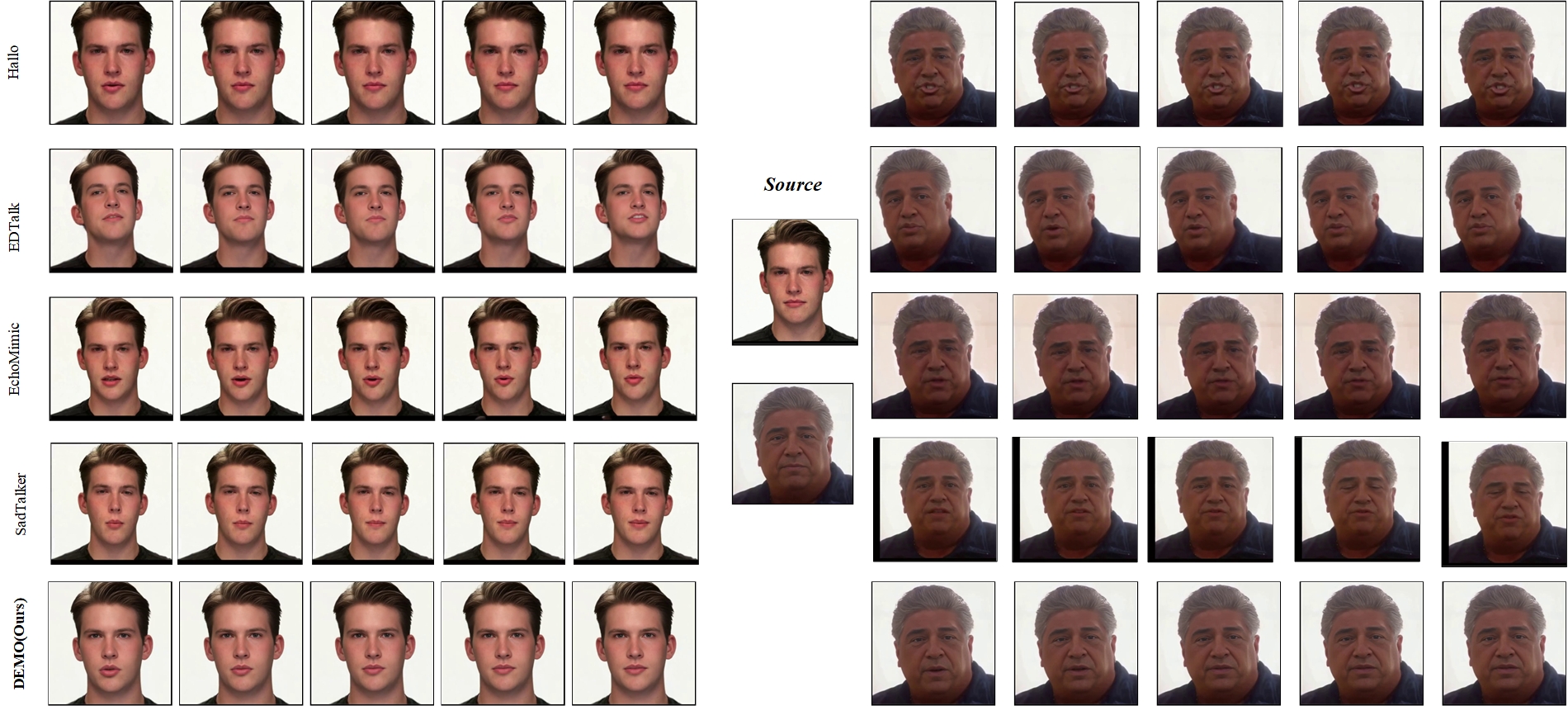}
	\caption{Qualitative comparison with existing approaches on RAVDESS/HDTF datasets.}
	\label{fig:qualitative comparison}
\end{figure}

\begin{figure}
	\centering
	\includegraphics[width=1\linewidth]{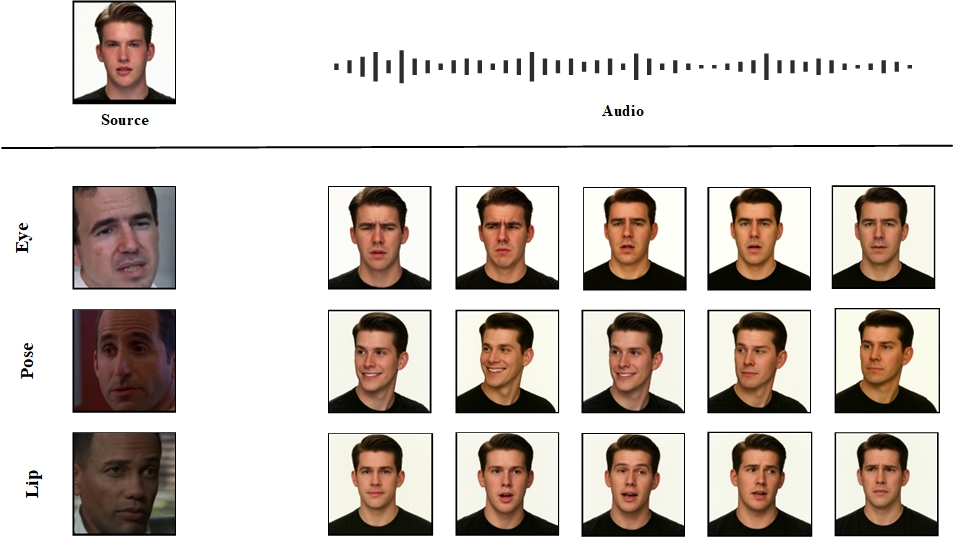}
	\caption{Fine-grained motion control with DEMO. Given a source image, a driving signal and a driving audio sequence, the framework varies only one motion factor (eye gaze, head pose, or lip movement) while keeping the others fixed.}
	\label{fig:illustration}
\end{figure}

\hspace{1em}We benchmark our method against a diverse set of state-of-the-art audio-driven talking-head models with publicly available implementations; for non-diffusion approaches, we include SadTalker \cite{zhang2023sadtalker} and EDTalk \cite{tan2024edtalk}; for diffusion-based methods, we evaluate against Hallo \cite{xu2024hallo}, and EchoMimic \cite{chen2025echomimic}. As shown in Fig.~\ref{fig:qualitative comparison}, Fig.~\ref{fig:illustration} and Table~\ref{tab:quantitative-comparison}, DEMO consistently outperforms these methods in both quantitative metrics and visual quality across the evaluation datasets.

\vspace{\baselineskip}
\textit{C. Ablation Study}

\textit{\textbf{1) Ablation on Fine-Grained Controllable Motion Encoder:}} To evaluate the contribution of Fine-Grained Controllable Motion Encoder (FCME), we replace it with a standard VAE and conduct driving experiments. As shown in Table~\ref{tab:ablation}, applying decorrelation strategies notably reduces FID and FVD scores, indicating improved factor disentanglement. Combining both strategies yields the largest gains. In particular, FCME enhances the separation of expression and lip dynamics, enabling more accurate and controllable motion synthesis.

\textit{\textbf{2) Ablation on Flow Matching:}} We further compare flow matching with a diffusion-based counterpart by adopting our vector predictor architecture as the denoising network. For fairness, we follow the diffusion training configuration of VASA-1 as an indirect reference. Results show that both approaches achieve comparable image fidelity (FID/FVD). However, flow matching delivers clear advantages in lip synchronization, evidenced by lower LSE-D and P-FID scores. This gain arises from the disentangled motion latent representation combined with OT-based flow matching, which together yield superior lip-sync alignment and natural head-motion dynamics.

\section{CONCLUSION}
\label{sec:typestyle}
\hspace{1em}In this paper, we propose DEMO, an audio-driven talking-head video generation framework that enables fine-grained and disentangled control of lip motion, head pose, and eye gaze. DEMO constructs a structured motion latent space with a motion auto-encoder, where individual facial motion factors are independently represented. Building on this representation, OT-based flow matching with a transformer predictor generates temporally coherent motion trajectories conditioned on audio. DEMO achieves state-of-the-art results, excelling in both perceptual quality (FID 94.05, CSIM 0.704) and lip–audio synchronization (P-FID 0.587, LSE-D 238.58). Extensive experiments across multiple benchmarks show that DEMO consistently surpasses existing methods in video realism, lip–audio synchronization, and motion fidelity. Our analysis demonstrates that disentangling motion factors and modeling flow-based trajectories significantly improve controllability, expressiveness, and temporal consistency, establishing a strong paradigm for high-fidelity, controllable talking-head synthesis and supporting realistic applications in virtual communication, film production, and interactive media.

{\footnotesize
\bibliographystyle{IEEEbib}
\bibliography{refs}
}

\end{document}